\pdfoutput=1

\documentclass[11pt]{article}

\usepackage{acl}

\usepackage{times}
\usepackage{blindtext}
\usepackage{latexsym}
\usepackage{amsmath}
\usepackage{amssymb}
\usepackage{dsfont}
\usepackage{graphicx}
\usepackage{subcaption}
\usepackage{threeparttable}
\usepackage{algorithm}
\usepackage{algorithmic}
\usepackage{multirow}
\usepackage{booktabs}
\usepackage{graphicx}
\usepackage{adjustbox}

\usepackage[T1]{fontenc}

\usepackage[utf8]{inputenc}

\usepackage{microtype}

\usepackage{inconsolata}

\usepackage{graphicx}
\usepackage{ stmaryrd }

\title{NeuroMax: Enhancing Neural Topic Modeling via Maximizing Mutual Information and Group Topic Regularization}

\author{
 \textbf{Duy-Tung Pham\textsuperscript{1, 2}\footnotemark[1]},
 \textbf{Thien Trang Nguyen Vu\textsuperscript{3}\footnotemark[1]},
 \textbf{Tung Nguyen\textsuperscript{1}\footnotemark[1]},
 \textbf{Linh Van Ngo\textsuperscript{1}\footnotemark[2]},
\\
 \textbf{Duc Anh Nguyen\textsuperscript{1}},
 \textbf{Thien Huu Nguyen\textsuperscript{4}},
\\
 \textsuperscript{1} Hanoi University of Science and Technology, Vietnam,
 \textsuperscript{2} FPT Software AI Center, Vietnam, \\
 \textsuperscript{3} VinAI Research, Vietnam,
 \textsuperscript{4} University of Oregon, USA,
}

\begin{document}

\maketitle

\renewcommand{\thefootnote}{\fnsymbol{footnote}} 
\footnotetext[1]{Equally contributed.} 
\footnotetext[2]{Corresponding author: \href{mailto:email@domain}{linhnv@soict.hust.edu.vn}} 
\renewcommand*{\thefootnote}{\arabic{footnote}}

\begin{abstract}
Recent advances in neural topic models have concentrated on two primary directions: the integration of the inference network (encoder) with a pre-trained language model (PLM) and the modeling of the relationship between words and topics in the generative model (decoder). However, the use of large PLMs significantly increases inference costs, making them less practical for situations requiring low inference times. Furthermore, it is crucial to simultaneously model the relationships between topics and words as well as the interrelationships among topics themselves. In this work, we propose a novel framework called NeuroMax (\textbf{Neur}al T\textbf{o}pic Model with \textbf{Max}imizing Mutual Information with Pretrained Language Model and Group Topic Regularization) to address these challenges. NeuroMax maximizes the mutual information between the topic representation obtained from the encoder in neural topic models and the representation derived from the PLM. Additionally, NeuroMax employs optimal transport to learn the relationships between topics by analyzing how information is transported among them. Experimental results indicate that NeuroMax reduces inference time, generates more coherent topics and topic groups, and produces more representative document embeddings, thereby enhancing performance on downstream tasks.

\end{abstract}

\section{Introduction}

Topic modeling \cite{1999plsi, blei2003lda, 2006dtm, li_2015_grouptopic, srivastava2017prodlda, bach2020tps, zhao2020nstm} is a well-established task in natural language processing (NLP) that involves uncovering and extracting latent topics from extensive corpora, thereby facilitating the comprehension and organization of unstructured data \cite{2019TMreview}. Its diverse applications span across fields such as text mining \cite{2017_effective, 2022podcast}, bioinformatics \cite{2020bioin4}, and recommender systems \cite{2018collaborative} and streaming learning \cite{ha2019eliminating,nguyen2021boosting,tuan2020bag}. 

Neural topic models \cite{srivastava2017prodlda, wang2022wete, wu2023effective, wu2024traco, zhao2020nstm, dieng2020etm} extend traditional topic modeling methods by incorporating neural network structures, thereby enhancing scalability and efficiency. Similar to Variational Autoencoders (VAEs) \cite{kingma2013vae}, neural topic models typically consist of two main components: an encoder (inference network) and a decoder (generative network). Recent research has focused on improving these components, leading to overall advancements in model performance.

Regarding the encoder, several studies have proposed incorporating knowledge from pretrained language models \cite{han-etal-2023-utopic, bianchi-etal-2021-crosslingual} such as BERT \cite{devlin2019bert} and GPT \cite{gpt1}. These models, trained on vast amounts of text data, effectively capture linguistic patterns and contextual information. This rich information can serve as input for the encoder \cite{bianchi-etal-2021-ctm, han-etal-2023-utopic}, enhancing the topic models' ability to generate coherent topics. However, despite such an advantage, utilizing large pretrained models significantly increases inference costs, which limits their utility in scenarios requiring low inference time.

Concerning the decoder, a line of work has leveraged pretrained word embeddings to better capture the semantics of the vocabulary \cite{dieng2020etm, zhao2020nstm, xu-etal-2023-vontss, 2022hyperminer, nguyen4592178out,nguyen2022adaptive}. Recently, \cite{wu2023effective} decomposed the topic-word distribution matrix into word and topic embeddings, with the word embeddings initialized by pretrained knowledge. The topic-word distribution is then modeled as the softmax of the negative $L_2$ distance between corresponding embeddings. Additionally, their method employs clustering regularization to group word embeddings into clusters, with each cluster corresponding to a topic, thereby mitigating the topic collapse problem. While these approaches improve efficiency in modeling word-topic relationships, they lack adequate consideration for capturing semantic connections between topics, resulting in a topic embedding space that is difficult to interpret.

\begin{figure}
    \centering
    \includegraphics[width=0.5\textwidth]{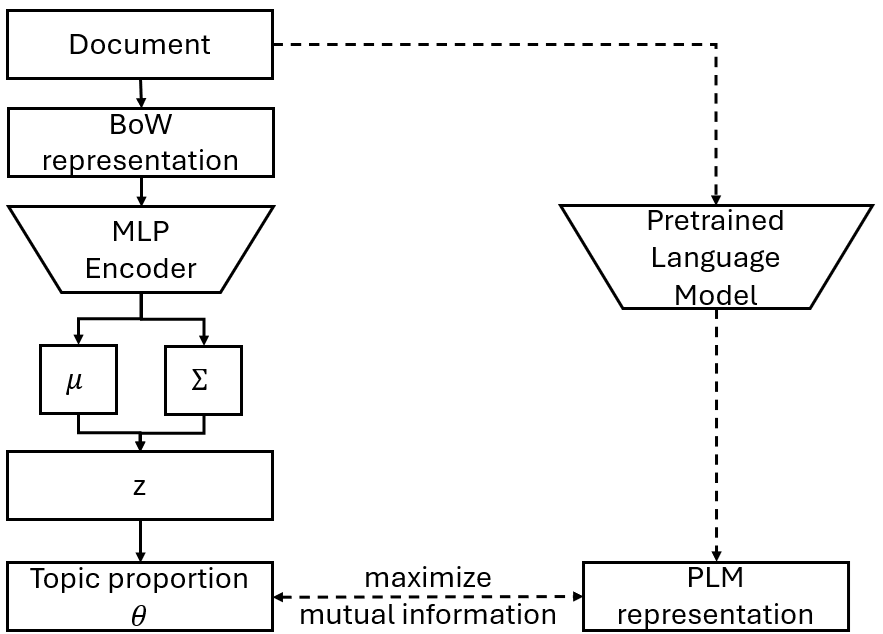}
    \caption{High-level architecture of our encoder. Dashed line represent the part of our model that could be excluded in inference time.}
    \label{fig:model}
\end{figure}

In response to these shortcomings, we propose a neural topic model framework that leverages the power of pretrained knowledge without incurring expensive inference costs and effectively captures semantic interrelationships at the topic level. First, for the encoder, we hypothesize that the topic proportions and embeddings derived from pretrained language models should exhibit similar representational characteristics. By maximizing the mutual information between these two variables, we integrate contextualized information from pretrained language models during training, thereby eliminating the need for pretrained components at the inference stage. Figure \ref{fig:model} presents the overall architecture of our proposed encoder, which operates without pretrained language models during inference. Second, inspired by \cite{van_assel2023snekhorn}, we employ optimal transport (OT) to model the relationships between topics. Specifically, in line with standard works \cite{wu2023effective}, we assume that each topic carries an equal amount of information, which is transported from one topic to another in a manner that preserves the total information within each topic. The learned transport plan elucidates the connections between the topics. Additionally, we assume that documents often encompass several closely related themes, naturally grouping topics into semantically related clusters. We enhance the group relationship of topics by imposing a regularization on the aforementioned transport plan based on predefined topic cluster relationships derived from topic clustering.

The rest of this paper is organized as follows: Section \ref{sec:relatedwork} lists some related works in topic modeling and neural topic modeling. Some background on neural topic models, mutual information maximization, and optimal transport is provided in \ref{sec:background}. Our proposed methodologies are introduced in \ref{sec:method}. Some experiments to illustrate the effectiveness of the proposed method are reported in \ref{sec:experiments}. Finally, our discussion and conclusion are given in Section \ref{sec:conclusion}.

\section{Related Work} \label{sec:relatedwork}

\textbf{Topic Models and Neural Topic Models.} Traditionally, generative probabilistic models such as Latent Dirichlet Allocation (LDA) \cite{blei2003lda} and Probabilistic Latent Semantic Indexing (PLSI) \cite{1999plsi} have been utilized for topic modeling. Numerous extensions of these models have been proposed to accommodate various assumptions and settings \cite{,duc2017keeping,nguyen2019infinite,van2022graph,li_2015_grouptopic, NGUYEN2022bsp, 2006dtm, bianchi-etal-2021-crosslingual}. Recent advancements have integrated topic models with Variational Autoencoders (VAE) \cite{kingma2013vae} to improve scalability and efficiency in the inference process \cite{srivastava2017prodlda, dieng2020etm, 2023infoctm, 2023neuralfocusdtm, wang2022wete, nguyen2024setwisemmo, bianchi-etal-2021-crosslingual, wu2024traco}. Due to the difficulty of sampling the Dirichlet prior using the reparameterization trick, a Laplacian approximation is utilized in \cite{srivastava2017prodlda}. An alternative method involves using rejection sampling variational inference for the Dirichlet prior \cite{2019dvae}.

Within VAE architecture, two lines of research aim at improving the two components of the model, the encoder and the decoder. In both directions, incorporating external knowledge like word embeddings has become a prevalent practice to enhance topic quality \cite{bach2020tps, dieng2020etm, bianchi-etal-2021-ctm, grootendorst2022bertopic, sia-etal-2020-tired-clustertm}. Regarding the decoder (or the reconstruction phase of topic modeling), \cite{dieng2020etm} proposed using word embeddings, such as Word2Vec \cite{mikolov2013word2vec} or GloVe \cite{pennington-etal-2014-glove}, to gain a better understanding of vocabulary semantics, thus creating topics with more semantically related words. Various variants of word embedding are considered; for example, \cite{xu-etal-2023-vontss} utilized spherical embeddings to improve clusterability, while \cite{2022hyperminer} employed word embeddings in hyperbolic space for topic taxonomy. In the embedding space, dependency between words and topics is modeled as a similarity function, such as dot product \cite{dieng2020etm, NGUYEN2022bsp}, cosine similarity \cite{zhao2020nstm}, or exponent of negative $L_2$ distance \cite{wu2023effective}. Additionally, \cite{wu2023effective} applied a clustering regularization technique to ensure that each topic embedding acts as the center of a distinct cluster of word embeddings, thereby mitigating the issue of topic collapse.

In terms of encoders, pretrained language models \cite{devlin2019bert, gpt1} are another type of external knowledge frequently incorporated. \cite{bianchi-etal-2021-ctm, bianchi-etal-2021-crosslingual} used contextualized document embedding from SBERT \cite{reimers-gurevych-2019-sbert} as input, capturing valuable information in complement to bag-of-word representation. \cite{han-etal-2023-utopic} further employed SBERT to generate term weights that are integrated into the reconstruction loss to filter out irrelevant words. These approaches lead to increased inference time for neural topic models, which poses challenges for real-time applications. An improvement that does not rely on external knowledge involves utilizing optimal transport distance to model the disparity between documents and topics \cite{wang2022wete, zhao2020nstm}.

Along with the development of pretrained language models, an alternate line of research in topic modeling that does not employ a VAE-like architecture directly group document's embedding to generate topics \cite{grootendorst2022bertopic, sia-etal-2020-tired-clustertm, zhang-etal-2022-better}. Although this approach is simpler and yields coherent topics, it is not trivial to infer the topic proportions for a document.

\textbf{Mutual Information Maximization.} Mutual information maximization has been extensively utilized in machine learning to develop representations that encapsulate the intrinsic structure of data \cite{oord2019infonce, hjelm2019deepinfomax}. For example, \cite{pmlr-v162-guo22ocm} employed this method to mitigate catastrophic forgetting in continual learning, while \cite{radford2021clip} leveraged mutual information to align text embeddings with image embeddings. In the field of topic modeling, mutual information maximization has been applied to align topics across different languages \cite{2023infoctm} and to derive meaningful document representations \cite{2021contrastiventm}.

\section{Background} \label{sec:background}

\textbf{Notations.} $\mathbf{X}=\{\mathbf{x}_i\}_{i=1}^{D}$ is a collection of D documents. $\mathbf{x}_{i\mathrm{BoW}}, \mathbf{x}_{iPLM}$ are the corresponding bag-of-words representation and pretrained language model embeddig of document $\mathbf{x}_i$. $V$ is the number of unique terms in our vocabulary. $K$ is the number of desired topics to find. $\beta=(\beta_1,\dots,\beta_K) \in \mathbb{R}^{V\times K}$ denotes the topic-word distribution matrix. $\mathbf{W}\in \mathbb{R}^{V\times L}, \mathbf{T}\in \mathbb{R}^{K\times L}$ correspond to the word embeddings and topic embeddings, respectively. $\theta_i$ is the topic proportion of document $\mathbf{x}_i$. $\mathds{1}_N$ is a vector of length N where every element is 1. $\llbracket n\rrbracket$ is the set of first $n$ integers $\{1, 2, \dots, n\}$. $\Delta^n$ is the probability simplex in $\mathbb{R}^n$: $\Delta^n = \{\theta \in \mathbb{R}^n \vert \theta_i \geq 0; \sum_{i=1}^n \theta_i = 1\}$. The inner product between two matrices $A$ and $B$ of the same size is represented as $\langle A, B \rangle = \sum_{i,j} A_{ij} B_{ij}$. The inner product between two vectors is defined similarly. $H(P) = -\langle P, \log P - 1 \rangle = - \sum_{i,j} P_{ij} (\log P_{ij} - 1)$ is the Shannon entropy of $P$. The KL divergence between $P$ and $Q$ is defined as $\mathrm{KL}(P \Vert Q) = \sum_{i,j} P_{ij} \log \left( \frac{P_{ij}}{Q_{ij}} \right) - 1$.

\label{section:background-ntm}

The objective of neural topic modeling is to identify $K$ latent topics within $\mathbf{X}$. Each topic is represented as a multinomial probability distribution over the $V$ vocabulary words, resulting in a topic-word distribution matrix $\beta \in \mathbb{R}^{V \times K}$. The matrix $\beta$ is then decomposed into two components: word embeddings and topic embeddings \cite{dieng2020etm, 2022hyperminer}. \cite{wu2023effective} defined the decomposition as follows:
\begin{equation*} \label{eq:beta_decomposition}
    \beta_{ij} = \frac{\exp \left(-\Vert \mathbf{w}_i - \mathbf{t}_j \Vert^2 / \tau \right)}{\sum_{j'=1}^K \exp \left(-\Vert \mathbf{w}_i - \mathbf{t}_{j'} \Vert^2 / \tau \right)}
\end{equation*}
where $\tau$ is a temperature hyperparameter. The word embeddings $\mathbf{W}$ are typically initialized using pre-trained word embeddings such as Word2Vec \cite{mikolov2013word2vec} or GloVe \cite{pennington-etal-2014-glove}. 

Another objective of neural topic models is to infer topic proportions for a document $\mathbf{x}_i$. \cite{srivastava2017prodlda, bianchi-etal-2021-ctm, dieng2020etm, wu2023effective} employ a VAE-like architecture. Specifically, the topic proportion $\theta$ depends on a latent variable $\mathbf{z}$, which conforms to a logistic-normal distribution characterized as $p(z)=\mathcal{L} \mathcal{N} \left( z \vert \mu_0, \Sigma_0 \right)$. When considering a document $\mathbf{x}_i$, its BoW representation, $\mathbf{x}_{i\mathrm{BoW}}$, is subjected to encoding through neural networks. These networks furnish the parameters of a normal distribution, with mean $\mu_i=h_\mu(\mathbf{x}_{i\mathrm{BoW}})$ and diagonal covariance matrix $\Sigma_i=\mathrm{diag}\left(h_\Sigma(\mathbf{x}_{i\mathrm{BoW}})\right)$. Leveraging the reparameterization trick \cite{kingma2013vae}, $z_i$ is sampled from the posterior distribution $q(z_i \vert \mathbf{x}_i)=\mathcal{N}\left(z_i \vert \mu_i, \Sigma_i \right)$ following $z_i=\mu_i +  \Sigma_i \varepsilon$, where $\varepsilon \sim \mathcal{N}(0,I)$. Subsequently, softmax function is applied to $z$, yielding topic proportion $\theta_i=\mathrm{softmax}(z_i)$. Following this, the Bag-of-Words representation is reconstructed with the topic-word distribution matrix $\beta$ from a multinomial distribution $\hat{\mathbf{x}}_{i\mathrm{BoW}} \sim \mathrm{Multi} \left(\mathrm{softmax} \left(\beta\theta_i \right) \right)$. This comprehensive process is instrumental in achieving our objective function for topic modeling, which consists of a reconstruction term and a regularization term as follows:
\begin{equation*} \label{eq:tm}
\begin{split}
\mathcal{L}_{\mathrm{TM}} = \frac{1}{D} \sum_{i=1}^{D} \Big[& - (\mathbf{x}_{i\mathrm{BoW}})^{\top} \log (\mathrm{softmax}(\mathbf{\beta} \theta_i)) \\ &+ \mathrm{KL} \left( q(z | \mathbf{x}_i) \| p(z) \right) \Big].
\end{split}
\end{equation*}

Other preliminary on mutual information maximization and entropic regularized optimal transport can be found in Appendix \ref{sec:appendix:preliminary}.

\section{Proposed Method} \label{sec:method}
We enhance both the inference network (encoder) and the generative model (decoder) of neural topic models through mutual information maximization and group topic regularization, respectively. The details will be presented in the subsequent subsections.

\subsection{Maximize Mutual Information with Pretrained Language Model}

We design an architecture that preserves the knowledge from a pretrained language model (PLM) in the encoder, even after the PLM is removed. Our approach is based on the assumption that the embeddings from the pretrained language model and the topic proportions should exhibit high mutual information. Specifically, let $\mathbf{X}_{\mathrm{PLM}}$ denote the distribution of the embeddings from the pretrained language model and $\Theta$ represent the distribution of topic proportions of the documents. The desired property can be achieved by maximizing the mutual information between these two random variables, $I(\mathbf{X}_{\mathrm{PLM}}; \Theta)$.

For tractability, we alternatively maximize its lower bound \cite{oord2019infonce}:
\begin{equation*}
\begin{split}
    I(\mathbf{X}_{\mathrm{PLM}}; \Theta) \ge &\log{B}\\ 
    &+ \frac{1}{D} \sum_{i=1}^D \log{\frac{e^{f(\theta_{i}, \mathbf{x}_{i\mathrm{PLM}})}}{\sum_{\theta'\in B_i} e^{f(\theta', \mathbf{x}_{i\mathrm{PLM}})}}}
\end{split}
\end{equation*}
where $B_i$ is a set containing sampled topic proportions of document $i$, including a positive example and negative examples for $\mathbf{x}_{i\mathrm{PLM}}$. $B_i$ is chosen to be the set of topic proportions of documents in the same batch as $\mathbf{x}_i$, and therefore has a size of $B$. The function $f(\theta, x_{\mathrm{PLM}})$ quantifies the similarity between the topic proportion $\theta$ and the PLM's embedding $x_{\mathrm{PLM}}$. Specifically, we use  $f(a,b)=\frac{\langle \phi_{\theta}(a), b \rangle}{\Vert \phi_{\theta}(a)\Vert \cdot \Vert b \Vert}$, where $\phi_{\theta}$ are learnable linear projections. As $B$ is chosen to be a constant, we therefore minimize the following InfoNCE loss:
\begin{equation*} \label{eq:Linfonce}
    \mathcal{L}_{\mathrm{InfoNCE}} = \frac{-1}{D} \sum_{i=1}^D \log{\frac{e^{f(\theta_{i}, \mathbf{x}_{i\mathrm{PLM}})}}{\sum_{\theta'\in B_i} e^{f(\theta', \mathbf{x}_{i\mathrm{PLM}})}}}
\end{equation*}

\subsection{Group Topic Regularization} \label{sec:method:gr}

We now introduce a new topic regularization based on optimal transport (OT) \cite{peyré2020computationalOT} for the decoder. Specifically, we assume that each topic contains an equal amount of information and conduct a process where information is transferred between topics based on their relationships, ensuring the total amount remains unchanged. This process helps us learn the relationship between topics. To make it easier to relate to the mass redistribution problem in OT \cite{peyré2020computationalOT}, we use the metaphor of $K$ topics as $K$ piles of soil, each with an equal mass of $\frac{1}{K}$. After transportation, the mass of each pile of soil remains $\frac{1}{K}$. The transportation cost between two topics is calculated based on the distance between them in the embedding space. The matrix $C$ represents the transportation costs for all pairs of topics. The optimal transport plan $Q$ reveals the relationships between topics.

Formally, let $C \in \mathbb{R}^{K\times K}$ be the cost matrix in Euclidean space for topic embeddings $\{\textbf{t}_1, \textbf{t}_2, \dots, \textbf{t}_K\}$. The transport plan $Q$ is the solution to the following optimization problem:
\begin{equation} \label{eq:ot-gr}
\begin{split}
 \text{minimize} & \quad \langle Q, C \rangle - \epsilon H(Q) \\
 \text{subject to} & \quad Q \in \mathbb{R}^{K \times K}, \\
& \quad Q \mathds{1}_K = Q^\top \mathds{1}_K = \frac{1}{K} \mathds{1}_K, \\
& \quad Q_{i,i} = 0 \ \forall i \in \llbracket K\rrbracket. \\
\end{split}
\end{equation}
The regularization term $\epsilon H(Q)$ encourages the matrix $Q$ to become dense, thereby facilitating the sharing of information across multiple topics \cite{blondel2018denseplan}. To focus on the interrelationships between different topics, the constraint $Q_{i,i} = 0$ is imposed. In practice, we ensure that $Q_{i,i}$ remains sufficiently small by setting $C_{i,i}$ to a large value. Subsequently, the Sinkhorn algorithm is employed to solve the optimization problem \cite{2013sinkhorn}.

Furthermore, given that documents are assumed to be delivered in groups with similar semantic meanings, it is assumed that topics also exhibit a cluster structure. To enforce this structure, we introduce a regularization term that aligns matrix $Q$ with matrix $P$, which encodes the shared information between grouped topics, as follows:
\begin{equation} \label{eq:gr}
    \mathcal{L}_{\mathrm{GR}} = \mathrm{KL}\left( P \Vert Q \right)
\end{equation}

We propose a method to construct the matrix $P \in \mathbb{R}^{K \times K}$ to represent the shared information between grouped topics aligning on $Q$. Our goal is to categorize the $K$ topic embeddings into clusters that reflect closely related semantic relationships. In the initial training phases, word embeddings display minimal deviation from their initialized states, thereby maintaining most of their semantic associations and effectively guiding the development of the topic embeddings. Leveraging this semantic information, we employ the KMeans clustering method \cite{macqueen1967kmean} to partition the $K$ topics into $G$ clusters. Subsequently, we establish the matrix $\hat{P}$ in the following manner:
\begin{equation*}
    \hat{P}_{ij} = \begin{cases}
    1 & \text{if topics $i, j$ are the same cluster} \\
    u & \parbox[t]{.6\textwidth}{otherwise}.
  \end{cases}
\end{equation*}
where the hyperparameter $0 < u < 1$ controls the ratio of shared information between topics within different groups compared to those within the same groups. We construct the final predefined matrix $P$ by normalizing $\hat{P}$ so that the elements in each row or column sum to $\frac{1}{K}$. The normalization process involves iteratively normalizing row-wise and projecting onto the space of symmetric matrices.

\subsection{Overall objective function}
Our inference process and topic modeling loss function follow the conventional neural topic model, as noted in Section \ref{section:background-ntm}. Additionally, inspired by \cite{wu2023effective}, we employ the Embedding Clustering Regularization regularizer to mitigate the topic collapsing problem:
\begin{equation} \label{eq:ecr}
\begin{split}
    &\mathcal{L}_{\mathrm{ECR}} = \sum_{i=1}^V \sum_{j=1}^K \Vert \mathbf{w}_i-\mathbf{t}_j \Vert^2  \pi^*_{ij}
\end{split}
\end{equation}
where $\pi^*$ is the solution of the following optimization problem:
\begin{equation} \label{eq:ot-ecr}
\begin{split}
    \text{minimize} & \  \langle C_{\mathrm{WT}}, \pi \rangle - \nu H(\pi) \\
    \text{s.t.} &  \ \pi \in \mathbb{R}^{V\times K} \\
     & \ \pi \mathds{1}_K = \frac{1}{V} \mathds{1}_V, \pi^T \mathds{1}_V = \frac{1}{K} \mathds{1}_K \\
\end{split}
\end{equation}
where $C_{\mathrm{WT}} \in \mathbb{R}^{V \times K}$ is the distance matrix between word embeddings and topic embeddings. $\pi^*$ is obtained using the Sinkhorn algorithm \cite{2013sinkhorn}. In summary, in addition to the topic model objective, we use three loss functions: $\mathcal{L}_{\mathrm{ECR}}$ to capture word-topic relations, $\mathcal{L}_{\mathrm{GR}}$ to regularize topic-topic relations, and $\mathcal{L}_{\mathrm{InfoNCE}}$ to enhance the encoder.

We can now finalize our training process as a two-stage approach. The first stage aims to produce the matrix $P$ for the group regularizer with the following objective function:
\begin{equation} \label{eq:stage1}
\begin{split}
    \mathcal{L}_{\mathrm{stage_1}} = \ &\mathcal{L}_{\mathrm{TM}} + \lambda_{\mathrm{ECR}} \mathcal{L}_{\mathrm{ECR}} \\
    & + \lambda_{\mathrm{InfoNCE}} \mathcal{L}_{\mathrm{InfoNCE}} 
\end{split}
\end{equation}
In practice, the first training stage requires only a few epochs to achieve effective topic groups. After obtaining $P$ as described in Section \ref{sec:method:gr}, we proceed to the second stage using the following loss function:
\begin{equation} \label{eq:overall}
\begin{split}
    \mathcal{L}_{\mathrm{stage_2}} = \ &\mathcal{L}_{\mathrm{TM}} + \lambda_{\mathrm{ECR}} \mathcal{L}_{\mathrm{ECR}} \\
    &+ \lambda_{\mathrm{GR}} \mathcal{L}_{\mathrm{GR}} + \lambda_{\mathrm{InfoNCE}} \mathcal{L}_{\mathrm{InfoNCE}} 
\end{split}
\end{equation}
where $\lambda_{\mathrm{GR}}, \lambda_{\mathrm{InfoNCE}}, \lambda_{\mathrm{ECR}}$ are weight hyperparameters. The full algorithm are described in Appendix \ref{sec:appendix:algo}.

\section{Experiments} \label{sec:experiments}

\begin{table*}[]
\centering
\begin{tabular}{  l  c  c  c c c  c  c  }
\hline
          & \multicolumn{3}{c}{20NG} & & \multicolumn{3}{c}{BBC}                            \\ \cline{2-4} \cline{6-8} 
          & $\mathrm{NPMI}$        & $\mathrm{NPMI-In}$        & $\mathrm{C_p}$            && $\mathrm{NPMI}$ & $\mathrm{NPMI-In}$ & $\mathrm{C_p}$      \\ \hline
LDA $\ddagger$       & 0.0057           & 0.0801         & 0.0727        && -0.0746   & -0.0199 & -0.0684 \\
ProdLDA $\ddagger$   & -0.0158          & -0.0610        & -0.0429       && 0.0001    & -0.0105 & 0.0647  \\
ETM $\ddagger$    & 0.0052           & 0.1219         & 0.0527        && -0.0212   & 0.0441  & 0.0829  \\
CTM $\ddagger$      & -0.0161          & \underline{0.1244}         & -0.1415       && 0.0436    & 0.0714  & 0.2543  \\
ClusterTM $\ddagger$  & 0.0135           & -0.2870        & 0.0160        && 0.0255    & 0.0656  & 0.0588  \\
BertTopic $\ddagger$ & 0.0609           & -0.0903        & 0.2318        && -0.0007   & 0.0943  & 0.0747  \\
UTopic $\ddagger$   & \textbf{0.1069}           & 0.1130         & \textbf{0.4850}        && \underline{0.0938}    & \underline{0.1256}  & \textbf{0.5388}  \\ \hline
NeuroMax      & \underline{0.0929}           & \textbf{0.1810}         & \underline{0.4543}        && \textbf{0.1288}    & \textbf{0.2174}  & \underline{0.5310}  \\ \hline
\end{tabular}
\caption{\centering Topic coherence measures, for models containing 10 topics. Bold values and underlined values represent the best and second-best results, respectively. $\ddagger$ Results resported in \cite{han-etal-2023-utopic}.}
\label{experiment:utopic10}
\end{table*}

\begin{table*}[]
\centering
\begin{tabular}{  l  c  c  c c c  c  c }
\hline
                    & \multicolumn{3}{c}{20NG}                            && \multicolumn{3}{c}{BBC}                             \\
\cline{2-4} \cline{6-8} 
          & $\mathrm{NPMI}$        & $\mathrm{NPMI-In}$        & $\mathrm{C_p}$            && $\mathrm{NPMI}$ & $\mathrm{NPMI-In}$ & $\mathrm{C_p}$      \\ \hline
LDA $\ddagger$                & -0.0056         & 0.0661          & 0.0719          && -0.0718         & -0.0205         & -0.0709         \\
ProdLDA $\ddagger$ & -0.0227         & -0.0083         & -0.0634         && 0.0084          & 0.0110          & 0.0569          \\
ETM $\ddagger$                & 0.0234          & 0.0927          & 0.1207          && -0.0333         & 0.0251          & 0.0416          \\
CTM $\ddagger$                & -0.0086         & \underline{0.1149}          & 0.0156          && 0.0289          & \underline{0.1109}          & 0.3254          \\
ClusterTM $\ddagger$          & 0.0154          & -0.2863         & 0.0082          && 0.0339          & 0.0990          & 0.0908          \\
BertTopic $\ddagger$          & 0.0322          & -0.0563         & 0.1515          && 0.0456          & 0.0762          & 0.2556          \\
UTopic $\ddagger$             & \textbf{0.0653}          & \textbf{0.1231}          & \textbf{0.3709}          && \underline{0.0708}          & 0.1018          & \underline{0.3925}          \\
\hline
NeuroMax                & \underline{0.0469}          & 0.0904          & \underline{0.3104}          && \textbf{0.0742}          & \textbf{0.1432}           & \textbf{0.3938}\\
\hline
\end{tabular}
\caption{\centering Topic coherence measures, for models containing 20 topics. Bold values and underlined values represent the best and second-best results, respectively. $\ddagger$ Results resported in \cite{han-etal-2023-utopic}.}
\label{experiment:utopic20}
\end{table*}

\subsection{Settings} \label{sec:experiments-settings}

\textbf{Datasets. } We employ 20 News Groups (20NG) \cite{1995_20news}, a popular benchmark for topic modeling, AGNews \cite{2015yahooagnews}, a corpus contains news articles from more than 2000 sources, IMDB \cite{maas-etal-2011-imdb}, a dataset of movie reviews, Yahoo Answers (Yahoo) \cite{2015yahooagnews} a dataset contains questions and answers from the Yahoo! Answer website, and BBC \cite{2006bbc} - a corpus from BBC news website in 2004 and 2005.

\textbf{Evaluation Metrics.} We follow the evaluation methodology proposed in \cite{wu2023effective} to assess both the quality of topics and the quality of document-topic distributions. Topic quality is evaluated using measurements of topic coherence and topic diversity. For topic coherence, we employ $\mathrm{C_V}$, $\mathrm{NPMI}$, and $\mathrm{C_p}$, which are established metrics in topic modeling known for their high correlation with human judgment \cite{2015topiccoherence}. These coherence measures are computed using a version of the Wikipedia corpus\footnote{\url{https://github.com/dice-group/Palmetto/}} as an external reference corpus. The $\mathrm{NPMI}$ measure is also computed using the training dataset as a reference dataset (denoted as $\mathrm{NPMI-In}$). For topic diversity, we use the proportion of unique words among the topic words. Document-topic distribution quality is evaluated using NMI and Purity \cite{manning2008informationretrieval} on the document clustering task.

\begin{table*}[!ht]

\centering
\begin{tabular}{lccccccccc}
\hline
       & \multicolumn{4}{c}{20NG}          &\multicolumn{1}{c}{}&\multicolumn{4}{c}{Yahoo} \\ \cline{2-5} \cline{7-10}
       & $\mathrm{C_V}$ & $\mathrm{TD}$  & $\mathrm{Purity}$ & $\mathrm{NMI}$   && $\mathrm{C_V}$   & $\mathrm{TD}$    & $\mathrm{Purity}$ & $\mathrm{NMI}$   \\ \hline
LDA \dag   & 0.385     & 0.655 & 0.367  & 0.364 && 0.359  & 0.843   & 0.288  & 0.144 \\
ETM \dag   & 0.375     & 0.704 & 0.347  & 0.319 && 0.354  & 0.719   & 0.405  & 0.192 \\
NSTM \dag  & 0.395     & 0.427 & 0.354  & 0.356 && 0.39   & 0.658   & 0.395  & 0.241 \\
WeTe \dag  & 0.383     & \underline{0.949} & 0.268  & 0.304 && 0.367  & 0.878   & 0.389  & 0.252 \\
ECRTM \dag & 0.431     & \textbf{0.964} & \underline{0.560}  & \underline{0.524} && \underline{0.405}  & \textbf{0.985}   & \underline{0.550}   & \underline{0.295} \\
Utopic & \textbf{0.508}     & 0.860 & 0.530  & 0.454 && \textbf{0.468}  & 0.788   & 0.473  & 0.244 \\ \hline
NeuroMax   & \underline{0.435}     & 0.912 & \textbf{0.623}  & \textbf{0.570} && 0.404  & \underline{0.979}   & \textbf{0.588}  & \textbf{0.331} \\ \hline
       & \multicolumn{4}{c}{IMDB}          &\multicolumn{1}{c}{}& \multicolumn{4}{c}{AGNews}      \\ \cline{2-5} \cline{7-10}
       & $\mathrm{C_V}$ & $\mathrm{TD}$  & $\mathrm{Purity}$ & $\mathrm{NMI}$   && $\mathrm{C_V}$   & $\mathrm{TD}$    & $\mathrm{Purity}$ & $\mathrm{NMI}$   \\ \hline
LDA \dag   & 0.347     & 0.788 & 0.614  & 0.041 && 0.364  & 0.864   & 0.64   & 0.193 \\
ETM \dag   & 0.346     & 0.557 & 0.66   & 0.038 && 0.364  & 0.819   & 0.679  & 0.224 \\
NSTM \dag  & 0.334     & 0.175 & 0.658  & 0.040  && 0.411  & 0.8773  & 0.7719 & 0.324 \\
WeTe \dag  & 0.368     & 0.931 & 0.587  & 0.031 && 0.383  & 0.945   & 0.641  & 0.268 \\
ECRTM \dag & 0.393     & \textbf{0.974} & \underline{0.694}  & \underline{0.058} && \underline{0.466}  & \textbf{0.961}   & \underline{0.802}  & \underline{0.367} \\
Utopic & \textbf{0.429}     & 0.554 & 0.550  & 0.005 && \textbf{0.545}  & 0.838   & 0.768  & 0.303 \\ \hline
NeuroMax   & \underline{0.402}     & \underline{0.936} & \textbf{0.709}  & \textbf{0.061} && 0.385  & \underline{0.952}   & \textbf{0.804}  & \textbf{0.410} \\ \hline
\end{tabular}
\caption{\centering Topic quality, quantified by mean $\mathrm{C_V}$ and mean $\mathrm{TD}$, and document-topic quality, evaluated using mean $\mathrm{NMI}$ and mean $\mathrm{Purity}$, for a model containing 50 topics. Bold values and underlined values represent the best and second-best results, respectively. \dag Results reported in \cite{wu2023effective}. }
\label{experiment:ecrtm50}
\end{table*}

\begin{table*}[!ht]
\centering
\begin{tabular}{lccccccccc}
\hline
       & \multicolumn{4}{c}{20NG}          & \multicolumn{1}{c}{} &\multicolumn{4}{c}{Yahoo}                                                          \\ \cline{2-5} \cline{7-10}
       & $\mathrm{C_V}$ & $\mathrm{TD}$  & $\mathrm{Purity}$ & $\mathrm{NMI}$   && $\mathrm{C_V}$   & $\mathrm{TD}$    & $\mathrm{Purity}$ & $\mathrm{NMI}$   \\ \hline
LDA \dag   & 0.387     & 0.622 & 0.364  & 0.346 && 0.359                & 0.602                & 0.297                & 0.148                 \\
ETM \dag   & 0.369     & 0.573 & 0.394  & 0.339 && 0.353                & 0.624                & 0.428                & 0.208                 \\
NSTM \dag  & 0.391     & 0.473 & 0.383  & 0.363 && 0.387                & 0.659                & 0.405                & 0.242                 \\
WeTe \dag  & 0.352     & 0.742 & 0.338  & 0.348 && 0.353                & 0.544                & 0.444                & 0.269                 \\
ECRTM \dag & 0.405     & \underline{0.904} & \underline{0.555}  & \underline{0.494} && 0.389                & \underline{0.903}                & \underline{0.563}                & \underline{0.311}                 \\
Utopic & \textbf{0.523}     & 0.750 & 0.545  & 0.452 && \textbf{0.476}                & 0.612                & 0.549                & 0.305                 \\ \hline
NeuroMax   & \underline{0.412}     & \textbf{0.913} & \textbf{0.602}  & \textbf{0.516} && \underline{0.390}                & \textbf{0.922}                & \textbf{0.583}                & \textbf{0.329}                 \\ \hline
       & \multicolumn{4}{c}{IMDB}          & \multicolumn{1}{c}{} & \multicolumn{4}{c}{AG News}                                                               \\ \cline{2-5} \cline{7-10} 
       & $\mathrm{C_V}$ & $\mathrm{TD}$  & $\mathrm{Purity}$ & $\mathrm{NMI}$   && $\mathrm{C_V}$   & $\mathrm{TD}$    & $\mathrm{Purity}$ & $\mathrm{NMI}$   \\ \hline
LDA \dag    & 0.342     & 0.691 & 0.600    & 0.037 && 0.349                & 0.696                & 0.654                & 0.194                 \\
ETM \dag   & 0.341     & 0.371 & 0.648  & 0.037 && 0.371                & 0.773                & 0.674                & 0.204                 \\
NSTM \dag  & 0.34      & 0.255 & 0.659  & 0.039 && \underline{0.421}                & 0.832                & 0.764                & 0.359                 \\
WeTe \dag  & 0.293     & 0.638 & 0.589  & 0.025 && 0.363                & 0.827                & 0.699                & 0.271                 \\
ECRTM \dag & 0.373     & \textbf{0.887} & \underline{0.694}  & \underline{0.049} && 0.416                & \textbf{0.981}                & \underline{0.812}                & \textbf{0.428}                 \\
Utopic & \textbf{0.534}     & 0.656 & 0.553  & 0.004 && \textbf{0.548}                & 0.681                & 0.760                & 0.283                 \\ \hline
NeuroMax   & \underline{0.381}     & \underline{0.870} & \textbf{0.706}  & \textbf{0.059} && 0.406 & \underline{0.957} & \textbf{0.828} & \underline{0.389} \\ \hline
\end{tabular}
\caption{\centering Topic quality, quantified by mean $\mathrm{C_V}$ and mean $\mathrm{TD}$, and document-topic quality, evaluated using mean $\mathrm{NMI}$ and mean $\mathrm{Purity}$, for a model containing 100 topics. Bold values and underlined values represent the best and second-best results, respectively. \dag Results reported in \cite{wu2023effective}.}
\label{experiment:ecrtm100}
\end{table*}

\begin{table*}[]
\centering
\begin{tabular}{lccccccccc}
\hline
& \multicolumn{4}{c}{20NG}          & \multicolumn{1}{c}{} & \multicolumn{4}{c}{Yahoo}                                                               \\ \cline{2-5} \cline{7-10} 
& $\mathrm{C_V}$ & $\mathrm{TD}$  & $\mathrm{Purity}$ & $\mathrm{NMI}$ && $\mathrm{C_V}$ & $\mathrm{TD}$  & $\mathrm{Purity}$ & $\mathrm{NMI}$\\
\hline
ECRTM                  & 0.431                    & \textbf{0.964}                    & 0.560                      & 0.524 && 0.405	& \textbf{0.985}	 & 0.55	 & 0.295                   \\ \hline
NeuroMax                   & 0.435                    & 0.912                    & \textbf{0.623}                      & \textbf{0.570}             && 0.404	 & 0.979	& \textbf{0.588}	 & \textbf{0.331}       \\ 
w/o GR               & \textbf{0.437}                    & 0.940                    & 0.613                      & 0.554  && \textbf{0.410}	& 0.957	& 0.577	 & 0.324                  \\
w/o InfoNCE                    & \textbf{0.437}                    & 0.924                    & 0.595                      & 0.547  && 0.404	& 0.937	& 0.564	& 0.317                  \\ \hline
\end{tabular}
\caption{Ablation study on 20NG and Yahoo datasets.}
\label{experiment:ablation}
\end{table*}

\begin{table}[!ht]
\begin{adjustbox}{width=0.48\textwidth}
    
\begin{tabular}{lcccc}
\hline
              & 20NG & IMDB & AGNews & Yahoo\\ \hline
UTopic        & 44.46                   & 98.69                   & 39.48                     & 39.42                     \\
NeuroMax & \textbf{0.15}            & \textbf{0.24}           & \textbf{0.14}             & \textbf{0.15}             \\ \hline
\end{tabular}
\end{adjustbox}
\caption{\centering Inference time of UTopic and Ours for different datasets with 50 topics. Experiments conducted on a NVIDIA RTX 3060 GPU.}
\label{experiment:inferencetime}
\end{table}

\textbf{Baseline models.} The first line of topic models not incorporating pre-trained language models includes: \textbf{LDA}, a probabilistic topic model introduced by \cite{blei2003lda}, \textbf{ProdLDA} \cite{srivastava2017prodlda}, a variant of the LDA model that integrates Variational Autoencoders (VAEs), \textbf{ETM} \cite{dieng2020etm}, a neural topic model that incorporates word embeddings, \textbf{NSTM} \cite{zhao2020nstm}, which utilizes the Sinkhorn distance to model the discrepancy between document-word distributions and document-topic distributions, \textbf{WeTe} \cite{wang2022wete}, which alternately employs a conditional transport distance, and \textbf{ECRTM} \cite{wu2023effective}, which implements clustering regularization to improve topic coherence and distinctiveness. A series of topic models leveraging pre-trained language models includes two that employ a VAE-like architecture: \textbf{CTM} \cite{bianchi-etal-2021-ctm}, which utilizes contextualized embeddings as inputs to the neural topic model to capture richer semantic information, and \textbf{UTopic} \cite{han-etal-2023-utopic}, which integrates tf-idf into the reconstruction loss to filter out unrelated words in a topic. Additionally, there are two baseline models that do not use a VAE-like architecture: \textbf{ClusterTM} \cite{sia-etal-2020-tired-clustertm}, which clusters documents based on their contextual embeddings and utilizes term frequency (tf) to generate topic words, and \textbf{BertTopic} \cite{grootendorst2022bertopic}, which employs a class-based variation of tf-idf to generate topic representations.

\subsection{Topic Quality and Doc-Topic Distribution Quality}
\label{sec:ex-tq-dtdq}

We first conducted experiments to assess the efficacy of our approach in comparison to baseline methods utilizing a pre-trained language model. Specifically, we employed two datasets: 20 News Groups and BBC News. We adhere to the preprocessing procedures outlined by \cite{han-etal-2023-utopic}. Tables \ref{experiment:utopic10} and \ref{experiment:utopic20} present the results of three topic coherence measures for 10 and 20 topics, respectively. Our approach demonstrates performance comparable to that of UTopic and consistently outperforms other methods. This outcome is expected, as the integration of the PLM's knowledge does not directly maximize the mutual information between topic proportions and contextualized representations. Instead, it operates via a lower bound approximation, which leads to improved topic coherence compared to methods that do not rely on PLMs, but does not achieve comparable embedding quality to methods that directly utilize PLMs, resulting in suboptimal overall performance.

We subsequently conducted experiments to evaluate the overall topic quality and document-topic distribution quality across four datasets: 20NG, Yahoo, IMDB, and AG News. The bag-of-words representation was obtained following the preprocessing steps described in \cite{wu2023effective}, and the contextualized embeddings were obtained after removing newline characters. Tables \ref{experiment:ecrtm50} and \ref{experiment:ecrtm100} report the topic quality and document-topic distribution quality for 50 and 100 topics, respectively. We provide the descriptive statistic in Appendix \ref{appendix:stat}. UTopic, as discussed in the previous experiment, demonstrates superior topic coherence performance but performs worse in terms of document-topic distribution quality. Compared to other baselines, our method achieves comparable topic quality and superior clustering performance owing to the integrated contextualized information and group regularization, which enhance the distinguishability of topic groups. Moreover, we illustrate the topics and their relationships in Appendix \ref{sec:appendix:viz}.

\subsection{Ablation Study}

We conducted an ablation study on the 20NG dataset to analyze the impact of each component of our model on overall performance. Specifically, we iteratively remove the group regularizer and the InfoNCE loss, subsequently evaluating the model's performance. Table \ref{experiment:ablation} presents the results obtained. In terms of document-topic distribution quality, measured by $\mathrm{NMI}$ and $\mathrm{Purity}$, both components enhance the quality of the distribution. The model incorporating both components achieves the highest performance. Regarding topic quality, the model's performance remains competitive even with the removal of one component.

\subsection{Inference time}

We conducted experiments to measure the inference time of our model compared to UTopic, a model that utilizes contextualized embeddings as input. The results, presented in Table \ref{experiment:inferencetime}, demonstrate that our method achieves approximately 300 times faster inference while maintaining competitive performance, as discussed in Section \ref{sec:ex-tq-dtdq}. These results highlight that the method of maximizing mutual information can effectively address the issue of high inference costs associated with pre-trained language models with an acceptable performance tradeoff.

\section{Conclustion} \label{sec:conclusion}
In conclusion, this paper introduces NeuroMax, a novel framework designed to tackle critical challenges in neural topic modeling. By maximizing the mutual information between the topic representation obtained from the common encoder in neural topic models and the representation derived from the PLM and leveraging optimal transport to capture topic relationships, NeuroMax offers a comprehensive solution for improving topic modeling efficiency and quality. Our experimental results demonstrate that NeuroMax significantly reduces inference time and obtains more coherent topics and topic groups, thus enhancing document representation for downstream task effectiveness. With its innovative approach and promising results, NeuroMax represents a valuable contribution to the field of neural topic modeling.

\section*{Limitations} \label{sec:limitations}

Our proposed method has several limitations. First is the necessity to predefine the number of topics and groups as hyperparameters. This requirement is undesirable in real-world applications where the number of topics and topic groups are needed to be determined dynamically. A potential solution is to utilize the stick-breaking process, as demonstrated in \cite{chen-etal-2021-nTSNTM, 2020hitmvae}, which can automatically determine the number of topics necessary. Another limitation is the challenge of applying our method to other scenarios, particularly dynamic topic models, online learning, and streaming learning. Adapting our approach to effectively capture the relationships between topics in temporal data remains an area for future research.

\bibliography{custom}

\appendix

\section{Appendix}
\label{sec:appendix}

\subsection{Algorithm} \label{sec:appendix:algo}
The detail training algorithm for NeuroMax is presented in Algorithm \ref{algo:neuromax}.

\begin{algorithm*}[t]
    \caption{Learning NeuroMax}
	\label{algo:neuromax}
	\algsetup{linenosize=\small}
  	\small
\begin{algorithmic}
\REQUIRE{Document collection $\mathbf{X}$, pretrained language model $\mathrm{PLM}$, pretrained word embedding $\mathbf{W}_{\mathrm{pretrained}}$, number of topic $K$, total number of training epoch $N$, number of training epochs for the first stage $M$;}
\ENSURE{Encoder network's parameter $W_{\mathrm{enc}}$, linear projections' parameter $W_{\phi_{\theta}}$, word embedding $\mathbf{W}$, topic embedding $\mathbf{T}$, word-topic transport plan $\pi^*$, topic-topic transport plan $Q$;}
\STATE{Initialize $\mathbf{W}=\mathbf{W}_{pretrained}$}
\FOR  {$t = 1,2, \ldots, N$}

    \STATE{// \textit{Update parameters related to the encoder}} 
    \STATE{Update $W_{\mathrm{enc}}$ and $W_{\phi_{\theta}}$ using a gradient descent step based on the loss  $\mathcal{L} = \mathcal{L}_{\mathrm{TM}} + \lambda_{\mathrm{InfoNCE}} \mathcal{L}_{\mathrm{InfoNCE}}$}
     
    \STATE{// \textit{Update parameters related to the decoder}} 
    \IF{$t \leq M$} 
        \STATE{// \textit{Stage 1}} 
        \STATE {Calculate word-topic distance matrix $C_{WT}$}
        \STATE {Update $\pi^*$ as the solution of problem (\ref{eq:ot-ecr}) by Sinkhorn algorithm}
        
        \STATE {Update $\mathbf{T}, \mathbf{W}$ using a gradient descent step based on the loss $\mathcal{L} = \mathcal{L}_{\mathrm{TM}} + \lambda_{\mathrm{ECR}} \mathcal{L}_{\mathrm{ECR}} $}
        \IF{$t=M$}
            \STATE {Calculate matrix $P$}
        \ENDIF
    \ELSE
        \STATE{// \textit{Stage 2}} 
        \STATE {Calculate word-topic distance matrix $C_{WT}$ and topic-topic distance matrix $C$}
        \STATE {Update $Q$ and $\pi^*$  using  Sinkhorn algorithm to sovle the OT problems (\ref{eq:ot-gr}) and (\ref{eq:ot-ecr}) respectively}
    
        \STATE {Update $\mathbf{T}, \mathbf{W}$ using a gradient descent step based on $\mathcal{L} = \mathcal{L}_{\mathrm{TM}} + \lambda_{\mathrm{ECR}} \mathcal{L}_{\mathrm{ECR}} + \lambda_{\mathrm{GR}} \mathcal{L}_{\mathrm{GR}}$ }
    \ENDIF

\ENDFOR

\end{algorithmic}
\end{algorithm*}

\subsection{Implementation Details.}
\label{sec:appendix:impdetails}
Our implementation builds upon PyTorch \cite{ansel2024pytorch} and TopMost \cite{wu2023topmost, wu2023survey}, a publicly available toolkit for topic modeling. Palmetto \cite{2015topiccoherence} is used to quantify topic coherence.We employ the $\mathrm{allMiniLM-L6-v2}$ model \cite{reimers-gurevych-2019-sbert} as our pretrained language model. GloVe \cite{pennington-etal-2014-glove} serves as the initial word embedding. Following the architecture in \cite{wu2023effective}, we utilize the same encoder network, comprising a two-layer softplus-activated MLP and an additional layer for the mean and covariance of the latent variable. We also train our model for $N=500$ epochs with a batch size of 200, utilizing the Adam optimizer \cite{kingma2017adam} with a learning rate set to 0.002. The hyperparameter $u$ of the group regularizer is set to $\frac{1}{5}$, and the number of first-stage training epochs is set to 10. The weight hyperparameters are searched in ranges as follows:
\begin{itemize}
    \item $\lambda_{\mathrm{ECR}} \in \left[20, 40, 50, 60, 80, 100, 150, 200, 250\right]$
    \item $\lambda_{\mathrm{GR}} \in \left[1, 5, 10, 20, 50\right]$
    \item $\lambda_{\mathrm{InfoNCE}} \in \left[1, 10, 30, 50, 80, 100, 130, 150\right]$
\end{itemize}

\subsection{Preliminary}
\label{sec:appendix:preliminary}

\subsubsection{Mutual Infomation Maximization}
Let $X$ and $Y$ be two random variables. The mutual information between $X$ and $Y$, which quantifies the degree of dependence between the two variables, is defined as:
\begin{equation} \label{eq:mutual-information}
    I(X;Y) = \int_X \int_Y p(X,Y) \log \frac{p(X,Y)}{p(X)p(Y)} dx dy
\end{equation}

In general, directly maximizing this quantity is intractable. We resort to its lower bound for tractable maximization \cite{oord2019infonce}:
\begin{equation} \label{eq:infonce}
\begin{split}
    I(X;Y) & \geq \mathcal{L}_{\mathrm{InfoNCE}} \\
    &= \log{N} + \mathbb{E}_{p(x,y)} \left[ \log \frac{f(x,y)}{\sum_{y' \in B} f(x,y')} \right]
\end{split}
\end{equation}
where $ f(x,y) $ is a similarity score between $ x $ and $ y $, and $ B $ is the set containing one positive and $ N-1 $ negative examples. Intuitively, maximizing this lower bound encourages a data instance to have a high similarity score with its positive example and a low similarity score with its negative examples.

\subsubsection{Entropic Regularized Optimal Transport}
Let $u$ and $v$ be two discrete measures on the supports $\{x_1, x_2, \dots, x_n\} \subset \mathbb{R}^d$ and $\{y_1, y_2, \dots, y_m \} \subset \mathbb{R}^d$, respectively, with associated weights $(u_1, u_2, \dots, u_n)$ and $(v_1, v_2, \dots, v_m)$ satisfying $\sum_i u_i = \sum_j v_j$. Given a cost matrix $C \in \mathbb{R}^{n \times m}$, an optimal transport plan is defined as the solution to the following optimization problem \cite{peyré2020computationalOT}:

\begin{equation} \label{eq:ot}
\begin{split}
    \underset{P \in \mathbb{R}^{n \times m}}{\mathrm{minimize}} \ &\langle P, C \rangle \\
    \mathrm{subject \ to} \ & P \mathds{1} = u, P^\top \mathds{1} = v.
\end{split}
\end{equation}

The minimized objective function of the aforementioned problem can serve as a measure of the distance between two distributions. This distance, along with the optimal plan, can be approximated efficiently by solving a modified problem, specifically by introducing an entropic regularization term to \eqref{eq:ot} and employing iterative algorithms such as Sinkhorn algorithm \cite{2013sinkhorn}.

The entropic regularization encourages the optimal transport plan to be dense \cite{blondel2018denseplan}. Consequently, if $u \equiv v$, mass from a given atom is compelled to disperse to other atoms, with a higher proportion of mass allocated to nearer atoms. Therefore, the entropic regularized optimal transport plan can be utilized as a similarity measure for data points \cite{van_assel2023snekhorn}.

\subsection{Descriptive Statistic}
\label{appendix:stat}

In Tables \ref{tab:std50} and \ref{tab:std100}, we report the performance of the NeuroMax model, providing both the mean and standard deviation over five independent runs. For comparison, the results for the UTopic model are also included as a reference baseline.

\begin{table}[h!]
\centering
\caption{Comparison of UTopic and NeuroMax on 50 topics (mean $\pm$ std)}
\begin{adjustbox}{width=0.48\textwidth}
\begin{tabular}{lcccc}
\toprule
Dataset & Metric & UTopic & NeuroMax \\
\midrule
\multirow{4}{*}{20NG} 
& $\mathrm{C_V}$ & $0.508 \pm 0.006$ & $0.435 \pm 0.004$ \\
& $\mathrm{TD}$ & $0.860 \pm 0.032$ & $0.912 \pm 0.045$ \\
& $\mathrm{Purity}$ & $0.530 \pm 0.010$ & $0.623 \pm 0.022$ \\
& $\mathrm{NMI}$ & $0.454 \pm 0.003$ & $0.570 \pm 0.014$ \\
\midrule
\multirow{4}{*}{IMDB} 
& $\mathrm{C_V}$ & $0.429 \pm 0.014$ & $0.402 \pm 0.005$ \\
& $\mathrm{TD}$ & $0.554 \pm 0.045$ & $0.936 \pm 0.025$ \\
& $\mathrm{Purity}$ & $0.550 \pm 0.004$ & $0.709 \pm 0.001$ \\
& $\mathrm{NMI}$ & $0.005 \pm 0.001$ & $0.061 \pm 0.002$ \\
\midrule
\multirow{4}{*}{Yahoo} 
& $\mathrm{C_V}$ & $0.468 \pm 0.012$ & $0.404 \pm 0.002$ \\
& $\mathrm{TD}$ & $0.788 \pm 0.007$ & $0.979 \pm 0.003$ \\
& $\mathrm{Purity}$ & $0.473 \pm 0.009$ & $0.588 \pm 0.004$ \\
& $\mathrm{NMI}$ & $0.244 \pm 0.008$ & $0.331 \pm 0.002$ \\
\midrule
\multirow{4}{*}{AGNews} 
& $\mathrm{C_V}$ & $0.545 \pm 0.008$ & $0.385 \pm 0.007$ \\
& $\mathrm{TD}$ & $0.838 \pm 0.025$ & $0.952 \pm 0.026$ \\
& $\mathrm{Purity}$ & $0.768 \pm 0.018$ & $0.804 \pm 0.006$ \\
& $\mathrm{NMI}$ & $0.303 \pm 0.012$ & $0.410 \pm 0.007$ \\
\bottomrule
\end{tabular}
\end{adjustbox}
\label{tab:std50}
\end{table}

\begin{table}[h!]
\centering
\caption{Comparison of UTopic and NeuroMax on 100 topics (mean $\pm$ std)}
\begin{adjustbox}{width=0.48\textwidth}
\begin{tabular}{lcccc}
\toprule
Dataset & Metric & UTopic & NeuroMax \\
\midrule
\multirow{4}{*}{20NG} 
& $\mathrm{C_V}$ & $0.523 \pm 0.006$ & $0.412 \pm 0.003$ \\
& $\mathrm{TD}$ & $0.750 \pm 0.012$ & $0.913 \pm 0.002$ \\
& $\mathrm{Purity}$ & $0.545 \pm 0.006$ & $0.602 \pm 0.007$ \\
& $\mathrm{NMI}$ & $0.452 \pm 0.006$ & $0.516 \pm 0.005$ \\
\midrule
\multirow{4}{*}{IMDB} 
& $\mathrm{C_V}$ & $0.534 \pm 0.004$ & $0.381 \pm 0.005$ \\
& $\mathrm{TD}$ & $0.656 \pm 0.028$ & $0.870 \pm 0.027$ \\
& $\mathrm{Purity}$ & $0.553 \pm 0.003$ & $0.706 \pm 0.004$ \\
& $\mathrm{NMI}$ & $0.004 \pm 0.001$ & $0.059 \pm 0.003$ \\
\midrule
\multirow{4}{*}{Yahoo} 
& $\mathrm{C_V}$ & $0.476\pm0.013$ & $0.390 \pm 0.002$ \\
& $\mathrm{TD}$ & $0.612 \pm 0.044$ & $0.922 \pm 0.029$ \\
& $\mathrm{Purity}$ & $0.549 \pm 0.014$ & $0.583 \pm 0.005$ \\
& $\mathrm{NMI}$ & $0.305 \pm 0.010$ & $0.329 \pm 0.003$ \\
\midrule
\multirow{4}{*}{AGNews} 
& $\mathrm{C_V}$ & $0.548 \pm 0.003$ & $0.406 \pm 0.007$ \\
& $\mathrm{TD}$ & $0.681 \pm 0.021$ & $0.957 \pm 0.015$ \\
& $\mathrm{Purity}$ & $0.760 \pm 0.011$ & $0.828 \pm 0.010$ \\
& $\mathrm{NMI}$ & $0.283 \pm 0.011$ & $0.389 \pm 0.014$ \\
\bottomrule
\end{tabular}
\end{adjustbox}
\label{tab:std100}
\end{table}

\subsection{Topic visualization} \label{sec:appendix:viz}
\begin{figure*}[!ht]
    \centering
    \begin{subfigure}[b]{0.41\textwidth}
        \centering
        \includegraphics[width=1\textwidth]{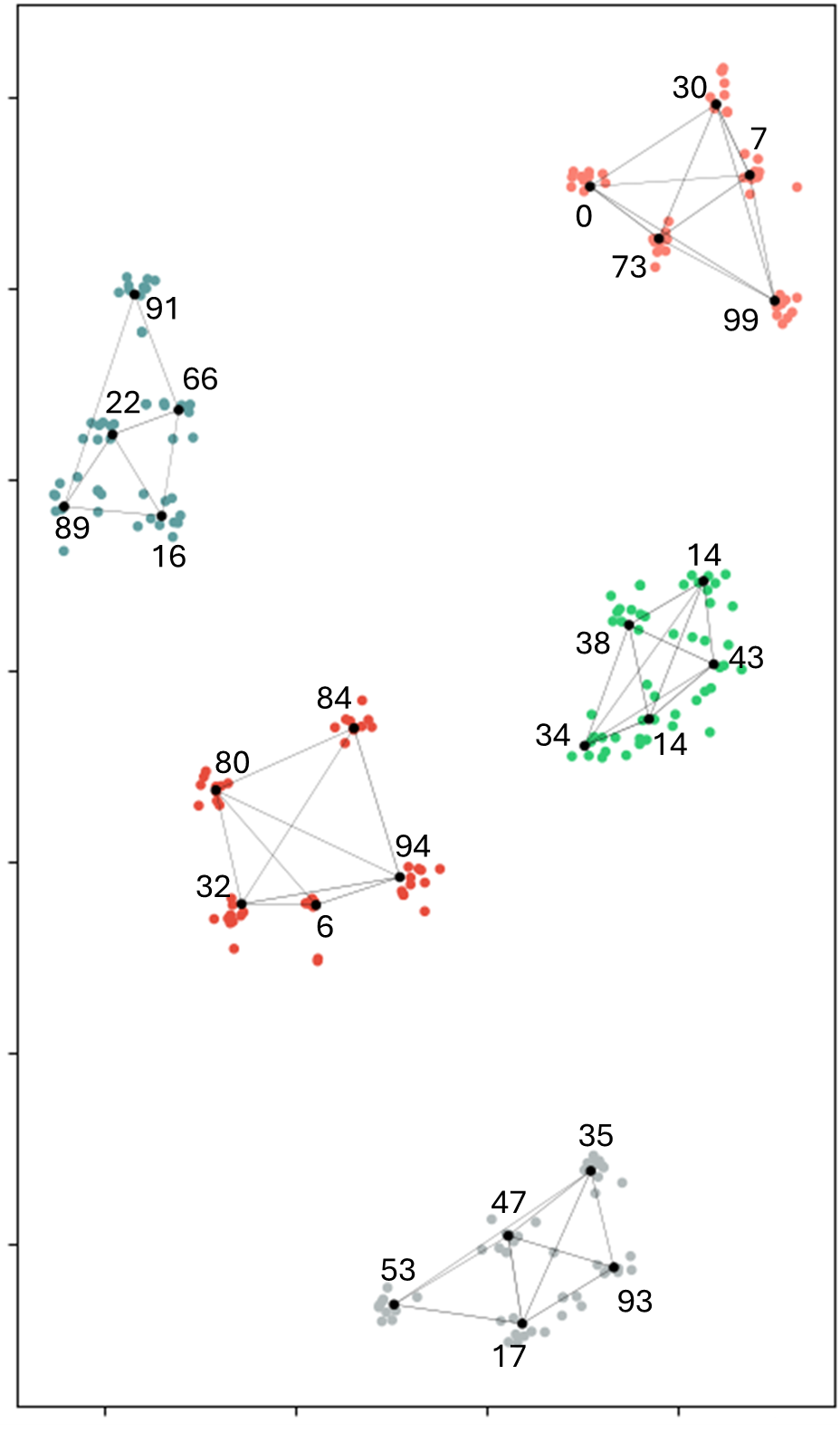}
    \end{subfigure}
    \hfill
    \begin{subfigure}[b]{0.57\textwidth}
        \centering
        \includegraphics[width=1\textwidth]{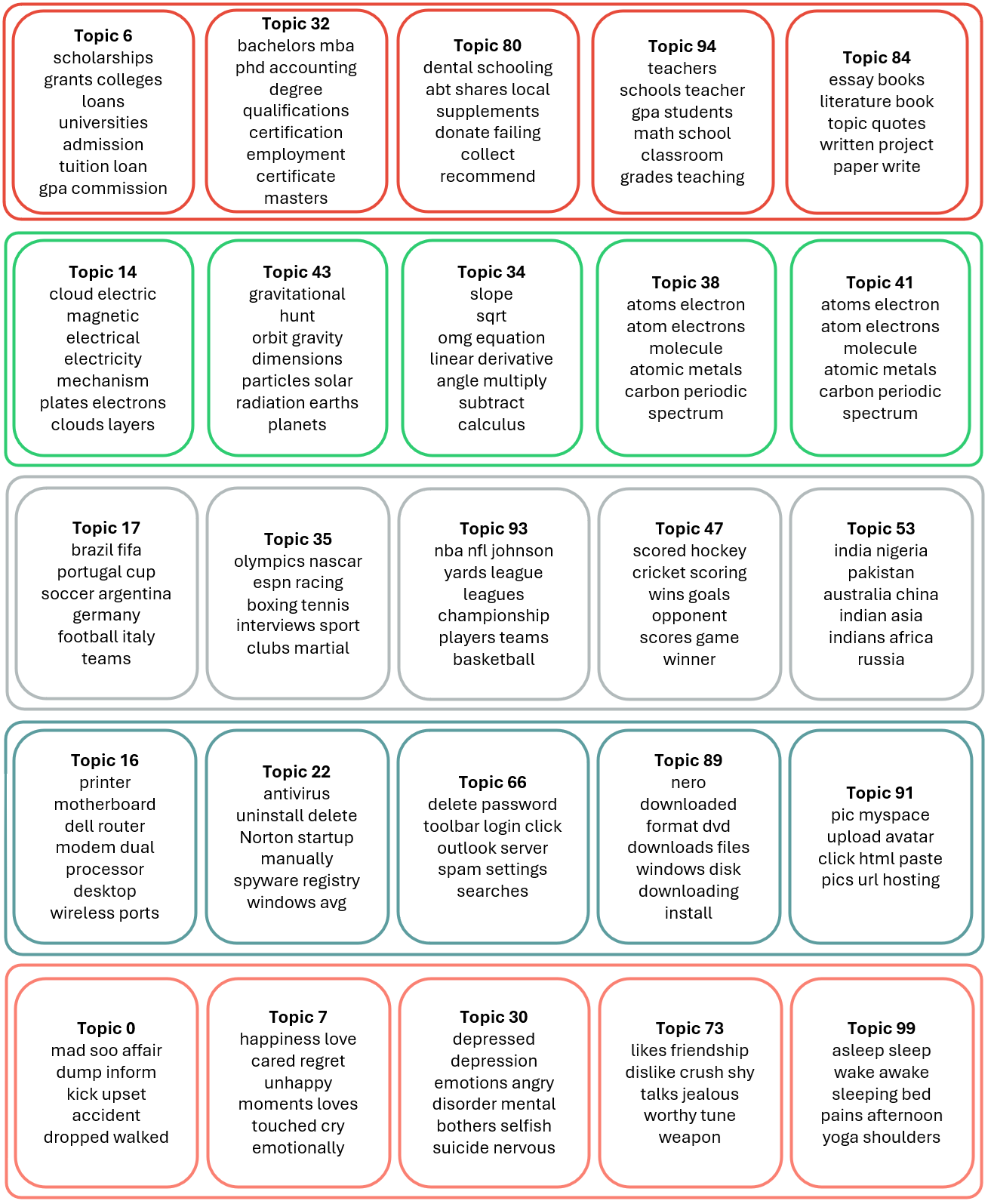}
    \end{subfigure}
    \caption{\centering (Left) t-SNE visualization of topics embeddings (black dots) and embeddings of their top 10 word (color dots). Word embeddings for topics within the same group share the same color. Pairs of topics with high information sharing scores are highlighted in gray. (Right) Corresponding top 10 words for each topic.}
    \label{fig:viz}
\end{figure*}

To assess the effectiveness of our grouping regularization, we visualized word and topic embeddings of 5 randomly selected groups, each comprising 5 randomly chosen topics, and displayed the corresponding topic words in Figure \ref{fig:viz}. We observed that topics within the same group tend to share more information (highlighted in gray) and share semantically similar words, while topics from different groups display distinct words and lower sharing scores. This highlights the efficacy of our group regularizer in generating closely embedded, semantically similar topics. Furthermore, the topics within the same group are not collapsing, thanks to the ECR regularizer.

\end{document}